\documentclass[a4paper]{article}
\usepackage{url}  
\usepackage{multirow}

\usepackage{INTERSPEECH2021}

\title{Three-Module Modeling For End-to-End Spoken Language Understanding Using
Pre-trained DNN-HMM-Based Acoustic-Phonetic Model}
\name{Nick J.C. Wang$^1$, Lu Wang$^1$, Yandan Sun$^1$, Haimei Kang$^1$, Dejun Zhang$^{1,2}$}
\address{
  $^1$Ping An Technology, China \\
  $^2$Yanshan University, Hebei, China}
\email{\{wangruizhang427,wanglu567,zhangdejun765\}@pingan.com.cn}

\begin{document}

\frenchspacing   
\maketitle
\begin{abstract}
  
  In spoken language understanding (SLU), what the user says is converted to
  his/her intent. 
  Recent work on end-to-end SLU has shown that accuracy can be improved via
  pre-training approaches.
  We revisit ideas presented by Lugosch et~al. 
  using speech pre-training and three-module modeling; however,
  to ease construction of the end-to-end SLU model,
  we use as our phoneme module
  an open-source acoustic-phonetic model from a DNN-HMM hybrid automatic speech
  recognition (ASR) system instead of training one from scratch.
  Hence we fine-tune on speech only for the word module, and we
  apply multi-target learning (MTL) on the word and intent modules to jointly
  optimize SLU performance. 
  MTL yields
  a relative reduction of 40\% in intent-classification error rates   
  (from 1.0\% to 0.6\%).                                    
  Note that 
  our three-module model is a streaming method.             
  The final outcome of the proposed three-module modeling approach  
  yields an intent accuracy of 99.4\% on FluentSpeech,                    
  an intent error rate reduction of 50\% compared to that of Lugosch et~al. 
  Although we focus on real-time streaming methods, we also list non-streaming
  methods for comparison. 

\end{abstract}
\noindent\textbf{Index Terms}: spoken language understanding (SLU), end-to-end
SLU, automatic speech recognition (ASR). 

\section{Introduction}

In conventional DNN-HMM hybrid speech recognition systems, acoustic models
use (sub)phonetic units generated by phonetic-context decision
trees~\cite{Povey_ASRU2011_2011, Gales2007}. 
Potentially large vocabulary systems can be built using pronunciation lexicons
that contain a relatively small number of such units. 
Such acoustic-phonetic modeling is prevalent in real-time commercial
ASR systems for the following two reasons.
It facilitates fast and accurate speech decoding via a weighted finite state transducer
(WFST) with an $N$-gram language model (LM), as well as a pronunciation
lexicon. 
Also, it accounts for the out-of-vocabulary (OOV) problem---words 
unseen during acoustic-model training---simply by adding them into the lexicon and
adding their training sentences into $N$-gram learning. 
However, OOVs are still a problem for
end-to-end models 
which use word-level units.                             
Far more speech data for training would be needed to account 
for unseen events when modeling word-level units. 
Moreover, decoding with advanced end-to-end models such as those
involving Transformers~\cite{Watanabe2017} can lead to increased 
decoding latency,                                       
as
such models require much longer input contexts.
One partial solution to these problems is to 
leverage the advantages of DNN-HMM-based acoustic-phonetic models. 
In this paper, we study the use of phonetic models and add a         
phoneme-to-word conversion module to build end-to-end models.


Traditionally, spoken language understanding systems are formulated as a pipelined
ASR component and text-based natural language understanding (NLU) component. 
The ASR component converts speech to word sequences which the NLU component then parses into
intents and slots. 
The trend toward end-to-end modeling has also entered the SLU domain. 
End-to-end SLU approaches map speech audio directly to the speaker intent
without explicitly producing a text transcript~\cite{Qian2017E2ESLU,
Serdyuk2018E2ESLU, Chen2018E2ESLU, Ghannay2018E2ESLU, Haghani2018E2ESLU,
Lugosch2019, Wang2020, Radfar2020, price2020end, Rongali2021}.
These thus eliminate the need for an independent ASR decoder and reduce overall
computational times in SLU decoding.  
In traditional pipelined systems, however, the hard decisions made by the ASR component
during decoding discard the acoustical distances from confusing words, which may contain the
true hypothesis, resulting in a final decision of the pipeline NLU
component that is suboptimal.
In contrast, an end-to-end SLU model can retain acoustical information
for all possible words in the middle layers, and hence allow these to compete with
each other via both acoustic and semantic scores in the final layers. 

Studies have been done on pre-training for end-to-end SLU modeling. 
Of studies which use FluentSpeech open-source data~\cite{Lugosch2019},
approaches without Transformer-based models~\cite{Lugosch2019, Wang2020, Radfar2020}
showed substantial improvements from pre-training, whereas those with Transformers
(or self-attention)~\cite{Radfar2020, price2020end, Rongali2021} showed no such improvements. 
Here we do not use Transformers, primarily because of their use of long input contexts and
increased decoding times, and secondarily because of their weakness when
using pre-training. 
However, we found that with a highly accurate Transformer ASR component, the pipelined NLU component's intent accuracy may outperform the other non-streaming end-to-end methods using Transformer. Our finding is presented in the final comparing table.  It may suggest room to further improve current end-to-end approaches.

For the proposed end-to-end SLU system, we use the three-module architecture 
from Lugosch~et~al.~\cite{Lugosch2019}. 
The scope and context length of each module are different,
ranging from phoneme to word to sentence.
Also, the context lengths range from short to long,
requiring optimization using different modeling approaches and data. 
The first module is for units at the phoneme or sub-phoneme level. 
Such units can compose all spoken words of different pronunciations.    
A conventional acoustic-phonetic model from a DNN-HMM hybrid ASR system may
already work well for this purpose. 
The second module is for units at the word or sub-word level. In a 
conventional DNN-HMM or GMM-HMM hybrid system, the decoder uses an explicit
mechanism with WFST and a pronunciation lexicon. 
In end-to-end models, this should be replaced by layers with phoneme-to-word
conversion effects, as in our second module. 
The third module is for units at the sentence level. 
In SLU, this converts word sequences to intents. 
For these three modules, we use pre-training approaches with 
out-of-domain speech data to learn the first and second modules
with phoneme and word targets. 

\section{Related Work}

\subsection{End-to-end SLU architecture of Lugosch et~al.}

Lugosch et~al.~\cite{Lugosch2019} use a three-module modeling
approach for end-to-end SLU that involves two major steps: phoneme and
word module training using speech data, and intent module training using SLU data.
They achieve 98.8\% intent classification accuracy on their
publicly released FluentSpeech data. 
Both phoneme and word modules are pre-trained using MTL
on out-of-domain speech data, that is, LibriSpeech~\cite{Panayotov2015}.
pre-training reduces their intent classification error by 65\%: from
3.4\% without pre-training to 1.2\% with pre-training.
We adopt their three-module architecture but use different methods for
pre-training and MTL. 
We replace the phoneme module with a well-trained, open-source speech
model, as described in the next section. 
Also, we use MTL on word and intent modules with SLU
data to jointly optimize SLU performance.

\subsection{Pre-trained Kaldi LibriSpeech model}\label{sec:chainmodel}

In this study we use the publicly available LibriSpeech Kaldi chain
model\footnote{\url{http://www.kaldi-asr.org/models/m13}} as the phonetic module.
This is a 16-layer factored time-delayed neural network (TDNN-F) model trained
with lattice-free maximum mutual information 
(LF-MMI) criteria~\cite{Povey2016} on 960~hours of LibriSpeech 
data~\cite{Panayotov2015} with a 3$\times$ speed perturbation.
%
A total of 6024 pdf-ids are generated via decision-tree clustering over
Gaussian mean vectors of triphone states~\cite{Nock1997}.
As input acoustic features, used are 40-dimensional mel-frequency
cepstral coefficients (MFCC) with i-vector speaker 
normalization~\cite{Senior2014}. 
In the experiments reported here, the parameters of this module were used
unmodified. 

\section{End-to-end SLU}

SLU maps a $T$-length speech feature sequence, $X =
\{\mathbf{x}_t \in \mathbb{R}^D| t=1, ..., T\}$, to a semantic intent $u$.
Here, $\mathbf{x}_t$ is a $D$-dimensional speech feature vector (e.g.,
MFCC) at frame~$t$, and $u$ is an intent.

SLU is mathematically formulated using Bayesian decision theory, where the most
probable intent, $\hat{u}$, 
is estimated among all possible intents, $U$, with the whole model $\Theta$ as
\begin{equation}
  \hat{u} = \mathop{\arg\max}\limits_{u \in U} p_\Theta(u|X).
  \label{eq_slu1}
\end{equation}
The main problem of SLU is thus obtaining the posterior distribution
$p_\Theta(u|X)$.
When learning this we attempt to find the optimal parameters, $\hat{\Theta}$, that minimize
the cross entropy (CE) between its posterior distribution and the data
distribution in the training set, $\mathfrak{D}$:
\begin{equation}
  \hat{\Theta} = \mathop{\arg\min}\limits_{\Theta} \sum_{(X,u) \in \mathfrak{D}}{ - \log p_\Theta(u|X)  }.
  \label{eq_slu1Train}
\end{equation}

Traditionally, the SLU decoding problem as illustrated by Eq.~(\ref{eq_slu1})
is factorized into two distributions---$p_{\Theta_W}(W|X)$ for ASR 
and $p_{\Theta_U}(u|W)$ for NLU---as 
\begin{equation}
\begin{aligned}
  \hat{u} 
  & = \mathop{\arg\max}\limits_{u \in U} \sum_{W \in V}{ \{ p_{\Theta_W}(W|X) \cdot p_{\Theta_U}(u|W) \} } \\ 
  & \approx \mathop{\arg\max}\limits_{u \in U} \max_{W \in \mbox{N-best(X)}}{ \{ p_{\Theta_W}(W|X) \cdot p_{\Theta_U}(u|W) \} },
  \label{eq_slu2a} 
\end{aligned}
\end{equation}
where the ASR module, $\Theta_W$, first decodes a speech feature sequence,
$X$, into a word sequence, $W = \{\mathbf{w}_i \in \mathbb{V}| i=1, ..., M\}$;
followed by the NLU module, $\Theta_U$, to decode the word sequence, $W$, into
an intent, $u$. 
Here, $V$ is the decoding vocabulary, and $\mathbf{w}_i$ is the
$i$-th word in the word sequence of length~$M$, which may differ from
the length of the speech sequence.

In the two-model approach, ASR and NLU are typically trained on different
types of data without joint optimization over end-to-end speech-to-intent data.
Decoding using the maximum over the $N$-best ASR sentences is a
fast way to replace summing probabilities over all probable word sequences,
$W$. Many pipeline systems streamline calculations by using only the $1$-best
ASR result to decode intents, as
\begin{equation}
\begin{aligned}
 \begin{cases}
 \hat{W}    & = \mathop{\arg\max}\limits_{W \in V }{ p_{\Theta_W}(W|X) } \vspace{0mm} \\ 
 \hat{u}  & = \mathop{\arg\max}\limits_{u \in U}{ p_{\Theta_U}(u|\hat{W}) }. 
 \end{cases}
  \label{eq_slu2b}
\end{aligned}
\end{equation}
Note that using only the $1$-best ASR result likely reduces
decoding accuracy. 

There are thus two advantages in using end-to-end SLU models:
end-to-end speech-to-intent probabilities can be jointly optimized, 
and SLU decoding is speeded up by reducing computational times in 
separate speech decoding phrases. 

\subsection{Three-module stepwise modeling}

\begin{table}[b]
  \caption{Modules in end-to-end SLU}
  \label{tab:3modules}
  \centering
  \begin{tabular}{lll}
    \toprule
    \textbf{Module} & \textbf{Symbol} & \textbf{Function} \\
    \midrule
    \textbf{Acoustic-phonetic}  & $\Theta_P$ & Waveforms to phones \\
    \textbf{Pronunciation}      & $\Theta_W$ & Phones to words \\
    \textbf{LU}                 & $\Theta_U$ & Words to intents \\
    \bottomrule
  \end{tabular}
\end{table}
As illustrated in Table~\ref{tab:3modules}, the SLU end-to-end models 
here are constructed using acoustic-phonetic, pronunciation, and 
language understanding (LU) modules,
with target units ranging from phoneme to word to sentence and context
lengths ranging from short to long. 
Such an end-to-end model can be expressed as a cascaded model:
${\left[\Theta_U,\Theta_W,\Theta_P\right]}$. 
Equation~(\ref{eq_slu1Train}) can thus be rewritten as
\begin{equation}
\begin{aligned}
  & {\left[\hat{\Theta}_U,\hat{\Theta}_W,\hat{\Theta}_P\right]} = \\ 
  & \mathop{\arg\min}\limits_{\left[\Theta_U,\Theta_W,\Theta_P\right]} \sum_{(X,u) \in \mathfrak{D}}{ - \log p_{\left[\Theta_U,\Theta_W,\Theta_P\right]}(u|X)  }.
  \label{eq_slu3Train}
\end{aligned}
\end{equation}
Stepwise learning is used to build the three modules one by one: first
$\hat{\Theta}_P$, then $\hat{\Theta}_W$, and last $\hat{\Theta}_U$. 

The first module, $\hat{\Theta}_P$, is constructed with alignments
of all the phone-level labels, $P= \{\mathbf{p}_t \in \mathbb{R}| t=1, ...,
T\}$, as
\begin{equation}
  {\hat{\Theta}_P} = \mathop{\arg\min}\limits_{\Theta_P} \sum_{(X,P) \in \mathfrak{D}}{ - \log p_{\Theta_P}(P|X)  }.
  \label{eq_slu3TrainP}
\end{equation}
Based on $\hat{\Theta}_P$, the second module, $\hat{\Theta}_W$, 
is constructed given transcription word sequences of length $L$,
$W = \{\mathbf{w}_l \in V| l=1, ..., L\}$, as
\begin{equation}
  {\hat{\Theta}_W} = \mathop{\arg\min}\limits_{\Theta_W} \sum_{(X,W) \in \mathfrak{D}}{ - \log p_{\left[\Theta_W,\hat{\Theta}_P\right]}(W|X)  }.
  \label{eq_slu3TrainW}
\end{equation}
Last, based on $\hat{\Theta}_P$ and $\hat{\Theta}_W$, 
given their annotated intent, $u$, the third module, $\hat{\Theta}_U$, is trained as
\begin{equation}
  {\hat{\Theta}_U} = \mathop{\arg\min}\limits_{\Theta_U} 
  \sum_{(X,u) \in \mathfrak{D}}{ - \log p_{\left[\Theta_U,\hat{\Theta}_W,\hat{\Theta}_P\right]}(u|X)  }.
  \label{eq_slu3TrainU}
\end{equation}

As described below, pre-training approaches using out-of-domain data are used
in the proposed method for three-module stepwise learning.
For example, the training of the first two modules uses out-of-domain speech
data without intent annotations. 
It is also possible to jointly optimize all modules with
in-domain SLU data. 
We conduct experiments on both configurations, as described below.

\subsubsection{Acoustic-phonetic (wave-to-phone) module, $\Theta_P$}

With the first module, $\Theta_P$, formulated as $\tilde{P} =
\mathit{Emb}_{\Theta_P}{(X)}$,
we convert a speech feature sequence, $X$, to a phoneme-level embedding vector
sequence, $\tilde{P}$, as the output of the last hidden layer. 
The resultant softmax values, $\mathrm{softmax}(\mathrm{linear}(\tilde{P}))$, are the
posterior probabilities used in Eq.~(\ref{eq_slu3TrainP}) to compute the
training error. 

Unlike Lugosch et~al.'s training of the speech model with MTL
using both phoneme and word targets,  
we adopt the publicly available non-end-to-end model for use in our
end-to-end SLU model, as mentioned in Section~\ref{sec:chainmodel}. 
The underlying expectation for the first module is to leverage the strengths
of conventional DNN-HMM-based acoustic-phonetic models. 
For instance, such phoneme-level modules
work around OOVs by using thousands of phoneme-level
units from which any word may be constructed, even 
words unseen during acoustic training;  
in addition, these modules achieve good accuracy with only hundreds 
of hours   
of speech training data. 
By contrast, end-to-end models using word-level units typically require far more
data and 
cannot recognize OOVs without complicated workarounds.  

\subsubsection{Pronunciation (phone-to-word) module, $\Theta_W$}

With the second module, $\Theta_W$, 
formulated as $\tilde{W} = \mathit{Emb}_{\Theta_W}{(\tilde{P})}$ given $\tilde{P}$,
we convert a phoneme-level embedding sequence, $\tilde{P}$, to a word-level
embedding sequence, $\tilde{W}$, as the output of the last hidden
layer. 
The resultant softmax values, $\mathrm{softmax}(\mathrm{linear}(\tilde{W}))$, are the
posterior probabilities used in Eq.~(\ref{eq_slu3TrainW}) to compute the
training error. 

Despite the advantages of the conventional DNN-HMM model described above,
to decode speech such a model requires hidden Markov models (HMMs), a pronunciation lexicon, a language model,
and a weighted finite state transducer (WFST). 
As the HMM mechanism is too simple to represent all pronunciation variations, 
directly using a neural network to map phone-level representations to
word-level representations is one way to improve performance.

Our second module for phoneme-to-word conversion is an LSTM recurrent neural 
network~\cite{Hochreiter1997lstm} trained using connectionist temporal classification
(CTC) loss~\cite{Graves2006, Graves2014} over byte pair encoding 
(BPE)~\cite{shibata1999byte} wordpieces from 
  BERT~\cite{devlin2018bert}.   

We use LibriSpeech for pre-training, followed by FluentSpeech for fine-tuning. 
By using the phoneme-level embedding vectors from the first module as input, 
less data may be required to converge when training the second module than
training directly using the speech data as input, because the underlying phoneme
embeddings provide more converged distributions than the raw speech data.

\subsubsection{LU (word-to-intent) module, $\Theta_U$}

With the last module, $\Theta_U$, formulated as an intent classifier
$\hat{u} = \mathop{\arg\max}\limits_{u \in U} p_{\Theta_U}(u|\tilde{W})$, we
convert a word-level embedding sequence, $\tilde{W}$, to an intent,
$\hat{u}$. 

This module handles sentence-scope information to make decisions. 
It is better to include the whole sentence as context and ignore the word
order if not important.
Currently, similar to Lugosch et~al.~\cite{Lugosch2019}, we use a recurrent neural network for
this with the output of the second module as its input, followed by a
linear-projection layer from the output of the last recurrent layer to all
classification targets before the softmax computation for the final
classification decisions. 
The model is fully trained on the small amount of target SLU data. 
In addition to stepwise training, we use MTL in order
to jointly optimize SLU, 
as described in the next section.



\subsection{Multi-target learning (MTL) for joint optimization}

As mentioned above, in all of our experiments we use an open-source model as the
first module, without further tuning on the target SLU data. 
The parameters of the second and third modules however, can be learned
jointly on the SLU data with multi-target learning (MTL) approach to optimize their SLU performance.
We modify Eq.~(\ref{eq_slu3TrainU}) as
\begin{equation}
\begin{aligned}
  & {\left[\hat{\Theta}_U,\hat{\Theta}_W\right]} = \\  
  & \mathop{\arg\min}\limits_{\left[\Theta_U,\Theta_W\right]} 
  \sum_{(X,W,u) \in \mathfrak{D}}{ - \log p_{\left[\Theta_U,\Theta_W,\hat{\Theta}_P\right]}(u,W|X)  },
  \label{eq_slu3TrainUW}
\end{aligned}
\end{equation}
where $\hat{\Theta}_P$ is fixed.
Here, the joint loss is computed as a weighted sum over logarithms of
intent posterior probabilities and word-sequence posterior probabilities with weight $\alpha$, as
\begin{equation}
\begin{aligned}
  & \log p_{\left[\Theta_U,\Theta_W,\hat{\Theta}_P\right]} (u,W|X) = \\ 
  & \alpha \log p_{\left[\Theta_U,\Theta_W,\hat{\Theta}_P\right]} (u|X) + (1 - \alpha) \log p_{\left[\Theta_W,\hat{\Theta}_P\right]} (W|X).
  \label{eq_LossMTL}
\end{aligned}
\end{equation}

\section{Experiments}

\subsection{Data sets}

We are grateful for the open-source FluentSpeech SLU 
data,\footnote{fluent.ai/research/fluent-speech-commands/}
a corpus of SLU data for spoken commands to smart homes or virtual
assistants, for instance, ``put on the music'' or ``turn up the heat in the kitchen''.
Each audio 
utterance   
is labeled with three slots: action, object, and location. 
A slot takes one of multiple values: for instance, the ``location'' slot can
take the values ``none'', ``kitchen'', ``bedroom'', or ``washroom''. 
In the paper, we follow Lugosch et~al.~\cite{Lugosch2019}, i.e., we refer to the
combination of slot values as the intent of the utterance, without
distinguishing between domain, intent, and slot prediction.
The dataset contains 31 unique intents and 30,043~utterances, or 19~hours of speech
in total, spoken by 97~speakers,
and is split into three parts: 23,132~utterances from 77~speakers for
training, 3,118~utterances from another 10~speakers for validation, and 
3,793~utterances from the remaining 10~speakers for testing.  
\begin{table}[ht]
  \caption{Speech datasets}
  \label{tab:datasets}
  \centering
  \begin{tabular}{lrrr}
    \toprule
    \textbf{Dataset} & \hspace{-3mm}\textbf{Speakers} & \textbf{Utterances}    & \textbf{Hours}                \\
    \midrule
    LibriSpeech Train      & 2,338    & -     & 960.7       \\
    FluentSpeech Train      & 77    & 23,132    & 14.7      \\
    FluentSpeech Valid      & 10    & 3,118     & 1.9       \\
    FluentSpeech Test       & 10    & 3,793     & 2.4       \\
    \bottomrule
  \end{tabular}
\vspace{-3mm}
\end{table}

We used another open-source data source, the LibriSpeech ASR 
corpus\footnote{\url{http://www.openslr.org/12}}~\cite{Panayotov2015}, to pre-train the first two modules of our
end-to-end SLU models.
This is a corpus of read speech derived from audiobooks, comprising approximately 1000 hours
of 16kHz read English speech. 
We used its 960 hours of speech in the training set to pre-train our
phone-to-word module. Table~\ref{tab:datasets} describes the datasets used in our
experiments.

\subsection{Pronunciation (phone-to-word) module modeling}

The second module converts (sub)phonetic units to (sub)word
units. We used BERT's 32K byte pair encoding (BPE) wordpiece 
vocabulary~\cite{devlin2018bert, shibata1999byte} as our word-level units 
to facilitate future text-based pre-training approaches. 
For this module we created a network with 4~layers of 768 UniLSTM units. 
Its input was the output of the above-mentioned first module of Kaldi
LibriSpeech model. 
Thus each frame of speech was first fed into the Kaldi TDNN-F LF-MMI
model to obtain its intermediate acoustic-phonetic embedding vector.
This module was trained over PyTorch platform using CTC loss~\cite{Graves2006, Graves2014} with a
linear projection of 768$\times$32K before applying softmax to compute the
posteriors, using the Adam optimization algorithm~\cite{Adam2014}
with an initial learning rate of $0.001$ (or $0.000125$ in the fine-tuning phase) and
$0.1$ dropout. 
The learning rate was cut in half when the validation set loss increased,
and learning was terminated when the loss did not decrease for three continuous
epochs. 
The second module was trained first on the LibriSpeech corpus, followed by 
fine-tuning with FluentSpeech SLU data for acoustic adaptation.

The wave-to-phone module and phone-to-word module together form an
end-to-end speech model which converts speech to words. 
Additionally, as mentioned before, our third module for LU can be fine-tuned
together with the second module to jointly optimize SLU performance. 
Table~\ref{tab:pipeline_exp} compares the ASR of the models in each modeling step. 
Clearly, fine-tuning reduces word error rates (WER) considerably and MTL yields a similar WER while improving its SLU intent accuracy. 

\subsection{End-to-end SLU Modeling}

The third intent module was added as a network with 2 layers of 768 UniLSTM
units, and trained after the above pronunciation module was ready. 
This module was trained using the PyTorch platform using cross-entropy (CE) loss with a linear projection of
768$\times$31 before applying softmax to compute the posteriors, and also
used Adam optimization. 
In addition to the above stepwise LU module training, the MTL LU modules with various $\alpha$ for the joint loss in Eq.~(\ref{eq_LossMTL}) are trained, too.
In Table~\ref{tab:pipeline_exp}, we compare the SLU intent accuracy of stepwise training and MTL with $\alpha = 0.5$: 
the latter outperforms the former.
Table~\ref{tab:e2e_exp} lists MTL experiments with different $\alpha$. MTL with $\alpha = 0.6$ yields the best intent accuracy: 99.4\%, which is currently the best result with the streaming approach. 
We also see in Table~\ref{tab:pipeline_exp} that
the intent accuracy of all end-to-end models is superior to that of the pipeline approach. 
Obvious superiority is observed over the `pre-trained' line, where its ASR WER is very high but its SLU intent accuracy for the end-to-end model is much better than for the pipeline approach. 
This demonstrates the superiority of end-to-end modeling. 

\begin{table}[tbh]
  \caption{ASR tests on FluentSpeech without language model, and two SLU tests (left: pipeline approach on ASR output, right: end-to-end approach)}
  \label{tab:pipeline_exp}
  \centering
  \begin{tabular}{lrrr}
    \toprule
    \multirow{2}{*}{\textbf{Pronunciation module}} & \hspace{-6mm}\multirow{2}{*}{\textbf{ASR WER}}   & \multicolumn{2}{c}{\textbf{SLU Intent acc.}} \\
    &  & \multicolumn{1}{c}{\textbf{Pipe.}} & \multicolumn{1}{c}{\textbf{E2E}} \\    \midrule
    LibriSpeech pre-trained      & 29.70\% & 62.8\% & 96.5\% \\
    +FluentSpeech fine-tuned   & 1.84\%    & 97.0\% & 99.0\% \\ 
    +FluentSpeech MTL 0.5        & 1.82\%    & 97.1\% & 99.2\% \\ 
    \bottomrule
  \end{tabular}
\end{table}

\begin{table}[tbh]
  \caption{SLU intent accuracy of tests on FluentSpeech with MTL end-to-end models trained with different weights}
  \label{tab:e2e_exp}
  \centering
  \begin{tabular}{cccccc}
    \toprule
    \textbf{MTL} & \multicolumn{5}{c}{\textbf{$\alpha$ in Eq.~(\ref{eq_LossMTL})}} \\
    \textbf{model} & 0.4 & 0.5 & 0.6 & 0.7 & 0.8 \\
    \midrule
    \textbf{E2E}     & 99.1\% & 99.2\% & 99.4\% & 99.3\% & 99.2\% \\ 
    \bottomrule
  \end{tabular}
\vspace{-3mm}
\end{table}

\subsection{Advanced non-streaming approaches}

Because SLU experiments using Transformer methods in Radfar et~al.~\cite{Radfar2020}
and Rongali et~al.~\cite{Rongali2021} do not benefit from pre-training, we are interested in
whether end-to-end SLU is as necessary as using advanced end-to-end models. 
We trained a CTC/attention Transformer ASR model using LibriSpeech following  ~\cite{Miao2020} with ESPNet\footnote{http://github.com/espnet/espnet} 
and experimented on pipeline decoding with FluentSpeech. 
Our Transformer has a 12-layer encoder and a 6-layer decoder, as well as a 2-layer CNN in front that reduces the frame rate to one fourth. 
This yields an intent accuracy of 99.6\% on FluentSpeech, outperforming the other methods. 
The SLU accuracies for these streaming and non-streaming methods 
are all listed in Table~\ref{tab:slu_comparison}.                             

\begin{table}[hb]
  \caption{Intent classification accuracy on FluentSpeech across various systems}
  \label{tab:slu_comparison}
  \centering
  \begin{tabular}{lr}
    \toprule
    \textbf{SLU approach} & \hspace{-16mm}\textbf{Intent acc.} \\
    \midrule
    \textbf{Streaming systems} &   \\
    \hspace{1mm} Pipeline: `+FluentSpeech.fine-tuned' + NLU     & 97.0\%  \\
    \hspace{1mm} E2E: Lugosch's et~al.~\cite{Lugosch2019}    & 98.8\%    \\
    \hspace{1mm} E2E: Our MTL 0.6 (w/ UniLSTM)    & 99.4\%    \\
    \textbf{Non-streaming systems} &  \\
    \hspace{1mm} E2E: Wang et al.~\cite{Wang2020}    & 99.0\%    \\
    \hspace{1mm} E2E: Price~\cite{price2020end}    & 99.5\%    \\
    \hspace{1mm} E2E: Rongali et~al.~\cite{Rongali2021}    & 99.5\%    \\
    \hspace{1mm} Pipeline: Our CTC/attn Transformer + NLU  & 99.6\%    \\ 
    \bottomrule
  \end{tabular}
\vspace{-3mm}
\end{table}

\section{Conclusion}

Using a publicly available phonetic module, we simplify the construction of
an end-to-end SLU model. Our pronunciation model for phoneme-to-word conversion is
pre-trained and fine-tuned for the SLU domain. We jointly optimize the LU module for
word-to-intent conversion with the
above phone-to-word module using MTL, yielding the superior
streaming-method intent accuracy of 99.4\% on FluentSpeech. A CTC/attention
Transformer model including long contexts results in the overall best intent
accuracy, 99.6\%, even with the pipeline decoding approach. This suggests           
end-to-end SLU still has room to improve.                                                

\bibliographystyle{IEEEtran}


\begin{thebibliography}{9}

\bibitem[1]{Lugosch2019}
L.\ Lugosch and M.\ Ravanelli and P.\ Ignoto and V.\ S.\ Tomar and Y.\ Bengio, 
  ``{Speech model pre-training for end-to-end spoken language understanding},''
  \textit{Proc. Interspeech}, pp. 814--818, 2019.
\bibitem[2]{Povey_ASRU2011_2011}
D.\ Povey and A.\ Ghoshal and G.\ Boulianne and L.\ Burget and O.\ Glembek and N.\ Goel,
  M.\ Hannemann and P.\ Motlicek and Y.\ Qian and P.\ Schwarz and J.\ Silovsky and G.\ Stemmer and
  K.\ Vesely, ``{The Kaldi speech recognition toolkit},'' \textit{IEEE 2011
  Workshop on Automatic Speech Recognition and Understanding (ASRU)}, pp.
  100--104, 2011.
\bibitem[3]{Gales2007}
M.\ Gales and S.\ Young, ``{The application of hidden Markov models in speech
  recognition},'' \textit{Found. Trends Signal Process.}, vol.\ 1, no.\ 3, p.
  195–304, 2007.
\bibitem[4]{Watanabe2017}
S.\ Watanabe and T.\ Hori and S.\ Kim and J.\ R. Hershey and T.\ Hayashi, ``{Hybrid
  CTC/attention architecture for end-to-end speech recognition},'' \textit{IEEE
  Journal of Selected Topics in Signal Processing}, vol.\ 11, no.\ 8, pp.
  1240--1253, 2017.
\bibitem[5]{Qian2017E2ESLU}
Y.\ Qian and R.\ Ubale and V.\ Ramanaryanan and P.\ Lange and D.\ Suendermann-Oeft and K.\ Evanini,
  and E.\ Tsuprun, ``{Exploring ASR-free end-to-end modeling to improve spoken
  language understanding in a cloud-based dialog system},'' \textit{IEEE
  Automatic Speech Recognition and Understanding Workshop (ASRU)}, pp.
  569--576, 2017.
\bibitem[6]{Serdyuk2018E2ESLU}
D.\ Serdyuk and Y.\ Wang and C.\ Fuegen and A.\ Kumar and B.\ Liu and Y.\ Bengio, ``{Towards
  end-to-end spoken language understanding},'' \textit{IEEE International
  Conference on Acoustics, Speech and Signal Processing (ICASSP)}, pp.
  5754--5758, 2018.
\bibitem[7]{Chen2018E2ESLU}
Y.-P. Chen and R.\ Price and S.\ Bangalore, ``{Spoken language understanding
  without speech recognition},'' \textit{IEEE International Conference on
  Acoustics, Speech and Signal Processing (ICASSP)}, pp. 6189--6193, 2018.
\bibitem[8]{Ghannay2018E2ESLU}
S.\ Ghannay and A.\ Caubriere and Y.\ Esteve and N.\ Camelin and E.\ Simonnet and A.\ Laurent and
  E.\ Morin, ``{End-to-end named entity and semantic concept extraction from
  speech},'' \textit{IEEE Spoken Language Technology Workshop (SLT)}, pp.
  692--699, 2018.
\bibitem[9]{Haghani2018E2ESLU}
P.\ Haghani and A.\ Narayanan and M.\ Bacchiani and G.\ Chuang and N.\ Gaur and P.\ Moreno,
  R.\ Prabhavalkar and Z.\ Qu and A.\ Waters, ``{From audio to semantics: Approaches
  to end-to-end spoken language understanding},'' \textit{IEEE Spoken Language
  Technology Workshop (SLT)}, pp. 720--726, 2018.
\bibitem[10]{Wang2020}
P.\ Wang and L.\ Wei and Y.\ Cao and J.\ Xie and Z.\ Nie, ``{Large-scale unsupervised
  pre-training for end-to-end spoken language understanding},'' \textit{IEEE
  International Conference on Acoustics, Speech and Signal Processing
  (ICASSP)}, 2020.
\bibitem[11]{Radfar2020}
M.\ Radfar and A.\ Mouchtaris and S.\ Kunzmann, ``{End-to-end neural transformer
  based spoken language understanding},'' \textit{Proceedings of the IEEE},
  vol.\ 77, no.\ 2, pp. 257--286, Feb. 2020.
\bibitem[12]{price2020end}
R.\ Price, ``{End-to-end spoken language understanding without matched language
  speech model pretraining data},'' \textit{IEEE International Conference on
  Acoustics, Speech and Signal Processing (ICASSP)}, pp. 7979--7983, 2020.
\bibitem[13]{Rongali2021}
S.\ Rongali and B.\ Liu and L.\ Cai and K.\ Arkoudas and C.\ Su and W.\ Hamza, ``{Exploring
  transfer learning for end-to-end spoken language understanding},'' \textit{The
  35th AAAI Conference on Artificial Intelligence}, 2021.
\bibitem[14]{Panayotov2015}
V.\ Panayotov and G.\ Chen and D.\ Povey and S.\ Khudanpur, ``{Librispeech: An ASR
  corpus based on public domain audio books},'' \textit{IEEE International
  Conference on Acoustics, Speech and Signal Processing (ICASSP)}, vol.\ 28,
  no.\ 4, pp. 357--366, 2015.
\bibitem[15]{Povey2016}
D.\ Povey, ``{Purely sequence-trained neural networks for ASR based on
  lattice-free MMI},'' \textit{IEEE Transactions on Acoustics, Speech and Signal
  Processing}, vol.\ 28, no.\ 4, pp. 357--366, Aug. 2016.
\bibitem[16]{Nock1997}
H.\ J. Nock and M.\ J. Gales and S.\ J. Young, ``{A comparative study of methods for
  phonetic decision-tree state clustering},'' \textit{Fifth European Conference
  on Speech Communication and Technology}, 1997.
\bibitem[17]{Senior2014}
A.\ Senior and I.\ Lopez-Moreno, ``{Improving DNN speaker independence with
  i-vector inputs},'' \textit{IEEE International Conference on Acoustics, Speech
  and Signal Processing (ICASSP)}, pp. 225--229, 2014.
\bibitem[18]{Hochreiter1997lstm}
S.\ Hochreiter and J.\ Schmidhuber, ``{Long short-term memory},'' \textit{Neural
  computation}, vol.\ 9, no.\ 8, pp. 1735--1780, 1997.
\bibitem[19]{Graves2006}
A.\ Graves and S.\ Fernandez and F.\ Gomez and J.\ Schmidhuber, ``{Connectionist
  temporal classification: Labelling unsegmented sequence data with recurrent
  neural networks},'' \textit{Proc. 23rd international conference on Machine
  learning}, pp. 369--376, 2006.
\bibitem[20]{Graves2014}
A.\ Graves and N.\ Jaitly, ``Towards end-to-end speech recognition with recurrent
  neural networks,'' \textit{International conference on machine learning}, pp.
  1764--1772, 2014.
\bibitem[21]{shibata1999byte}
Y.\ Shibata and T.\ Kida and S.\ Fukamachi and M.\ Takeda and A.\ Shinohara and T.\ Shinohara and
  S.\ Arikawa, ``Byte pair encoding: A text compression scheme that accelerates
  pattern matching,'' \textit{Technical Report DOI-TR-161, Department of
  Informatics, Kyushu University}, 1999.
\bibitem[22]{devlin2018bert}
J.\ Devlin and M.-W. Chang and K.\ Lee and K.\ Toutanova, ``{Bert: Pre-training of deep
  bidirectional transformers for language understanding},'' \textit{ArXiv}, vol.
  abs/1810.04805, 2018.
\bibitem[23]{Adam2014}
D.\ P. Kingma and J.\ L. Ba, ``{Adam: A method for stochastic optimization},''
  \textit{Conf. ICLR}, 2015.
\bibitem[24]{Miao2019}
H.\ Miao and G.\ Cheng and P.\ Zhang and T.\ Li and Y.\ Yan, ``{Online hybrid CTC/attention
  architecture for end-to-end speech recognition},'' \textit{INTERSPEECH}, pp.
  2623--2627, 2019.
\bibitem[25]{Miao2020}
H.\ Miao and G.\ Cheng and C.\ Gao and P.\ Zhang and Y.\ Yan, ``{Transformer-based online
  CTC/attention end-to-end speech recognition architecture},'' \textit{IEEE
  International Conference on Acoustics, Speech and Signal Processing
  (ICASSP)}, pp. 6084--6088, 2020.
\bibitem[26]{Vaswani2017}
A.\ Vaswani and N.\ Shazeer and N.\ Parmar and J.\ Uszkoreit and L.\ Jones and A.\ N. Gomez,
  L.\ Kaiser and I.\ Polosukhin, ``{Attention is all you need},'' \textit{Proc.
  31st International Conference on Neural Information Processing Systems
  (NIPS)}, 2017.
\end{thebibliography}

\end{document}